**Property-Level Flood Risk Assessment Using AI-Enabled Street-View Lowest Floor Elevation Extraction and ML Imputation Across Texas**

Xiangpeng Li[1*], Yu-Hsuan Ho [1], Sam D Brody[2], Ali Mostafavi[3]

[1] Ph.D. student. Urban Resilience.AI Lab, Zachry Department of Civil and Environmental Engineering, Texas A&M University, College Station, Texas, United States.

[2] Director, Institute for a Disaster Resilient Texas (IDRT) Regents Professor, Marine and Coastal Environmental Science, College of Marine Sciences & Maritime Studies, Texas A&M University at Galveston

[3] Professor. Urban Resilience.AI Lab, Zachry Department of Civil and Environmental Engineering, Texas A&M University, College Station, Texas, United States.

[*] Corresponding author: Xiangpeng Li, E-mail: xplli@tamu.edu.

## Abstract

Lowest floor elevation (LFE) is one of the most important yet least available inputs for flood loss estimation because flood depth can only be translated into structural damage when referenced to the elevation of the lowest occupied floor. This paper argues that AI-enabled analysis of street-view imagery, complemented by performance-gated machine-learning imputation, provides a viable pathway for generating building-specific elevation data at regional scale for flood risk assessment. We develop and apply a three-stage pipeline across 18 areas of interest (AOIs) in Texas that (1) extracts LFE and the height difference between street grade and the lowest floor (HDSL) from Google Street View imagery using the Elev-Vision framework, (2) imputes missing HDSL values with Random Forest and Gradient Boosting models trained on 16 terrain, hydrologic, geographic, and flood-exposure features, and (3) integrates the resulting elevation dataset with Fathom 1-in-100 year inundation surfaces and USACE depth-damage functions to estimate property-specific interior flood depth and expected loss. Across 12,241 residential structures, street-view imagery was available for 73.4% of parcels and direct LFE/HDSL extraction was successful for 49.0% (5,992 structures). Imputation was retained for 13 AOIs where cross-validated performance was defensible, with selected models achieving $R^2$_CV values from 0.159 to 0.974; five AOIs were explicitly excluded from prediction because performance was insufficient. The results show that street-view-based elevation mapping is not universally available for every property, but it is sufficiently scalable to materially improve regional flood-risk characterization by moving beyond hazard exposure to structure-level estimates of interior inundation and expected damage. Scientifically, the study advances LFE estimation from a pilot-scale proof of concept to a regional, end-to-end workflow. Practically, it offers a replicable framework for jurisdictions that lack comprehensive Elevation Certificates but need parcel-level information to support mitigation, planning, and flood-risk management.

## 1. Introduction

Floods are among the most consequential and economically damaging natural hazards globally. NFIP policyholders alone filed more than $72 billion in claims between 1990 and 2022(Tonn & Czajkowski, 2022), and these figures represent only a fraction of total flood-related losses that extend to uninsured properties and indirect economic disruption. Texas is among the nation's most acutely flood-exposed states, having experienced repeated catastrophic events including Tropical Storm Allison (2001), Hurricane Harvey (2017), and numerous intermediate events that have collectively caused hundreds of billions of dollars in cumulative losses (Li et al., 2026; Sebastian et al., 2021; Shultz & Galea, 2017). Effective flood risk management in such an environment demands not only accurate characterization of flood hazards (the spatial extent and depth of inundation) but equally rigorous characterization of structural vulnerability, and in particular the vertical relationship between a building's lowest habitable floor and the flood water surface.

Among structural characteristics, lowest floor elevation (LFE) is the single most important determinant of flood damage severity (Diaz et al., 2022a; Scawthorn et al., 2006). A vertical difference of as little as 0.30 m (one foot), the standard interval at which depth-damage functions are evaluated, can determine whether a structure experiences catastrophic interior flooding or escapes with negligible loss(Bodoque et al., 2016; Johnson, 2019). Despite this significance, regionally comprehensive LFE data is rarely available. The primary collection mechanism, the FEMA Elevation Certificate (EC), which must be completed and sealed by a licensed surveyor, engineer, or architect, is prohibitively costly and labor-intensive at scale, and EC coverage across the residential housing stock remains sparse even in communities with well-documented flood histories(Diaz et al., 2022b; Yildirim et al., 2022).

Regional flood risk assessment remains constrained by a fundamental imbalance between hazard data and vulnerability data (Ronco et al., 2014). Advances in hydraulic and compound flood modeling now make it possible to estimate flood extent and depth at high spatial resolution (Nederhoff et al., 2024), but the conversion of those hazard estimates into expected structural damage still depends on one of the least available building attributes: the elevation of the lowest occupied floor relative to floodwater. In practice, two houses exposed to the same modeled flood depth can experience very different outcomes because a few decimeters of floor elevation determine whether water remains below the threshold of entry or penetrates the structure interior (Gems et al., 2016). When LFE is missing, flood risk assessments are forced to rely on exposure metrics alone, which weakens the accuracy and usefulness of damage estimation for planning, mitigation, and risk communication.

This limitation is particularly consequential in Texas, where communities face repeated flood losses across highly diverse residential landscapes and where regionally complete survey-based elevation data are seldom available. Emerging work using street-view imagery suggests that AI can recover LFE-related information directly from building facades (Wu et al., 2021), but the literature has not yet established whether this promise extends beyond single-neighborhood pilots to geographically diverse, operational settings. Nor has it resolved how to handle the large fraction of properties for which direct estimation fails because imagery is missing or the features needed to infer floor height are obscured. These unresolved issues motivate the present study, which evaluates regional-scale deployability, introduces a performance-screened imputation strategy for missing HDSL, and tests how the resulting elevation data improve property-level and AOI-level flood-loss characterization.

Recent research has established that computer vision applied to freely available Google Street View (GSV) imagery offers a viable alternative to physical surveys (Chen et al., 2019a; Fan et al., 2023). Ning et al. (2022) provided the first proof-of-concept demonstration of object detection-based LFE estimation from street-view images, while Y. Ho et al. (2025; 2024) substantially extended this work through Elev-Vision, an image-segmentation approach that processes equirectangular GSV panoramas directly, achieves a mean absolute error (MAE) of 0.190 m, and additionally derives HDSL estimates critical for flood damage assessment. Both studies, however, were conducted at the neighborhood scale, with validation samples numbering in the hundreds. Whether the Elev-Vision framework maintains its accuracy and practical utility when deployed across tens of thousands of structures spanning the heterogeneous landscapes of a major metropolitan region, and whether machine learning can reliably fill the inevitable gaps in imagery coverage, are questions that have not been addressed.

The literature has made substantial progress in modeling flood extent and depth, but much less progress in characterizing the building-specific elevation information needed to translate exterior inundation into interior damage. Lowest floor elevation is central to that translation, yet regionally complete LFE data remain rare because the dominant data source—survey-based Elevation Certificates—is expensive, sporadic, and impractical to assemble at large scale. Recent studies using street-view imagery and computer vision demonstrate that automated LFE estimation is feasible, but they have largely remained proof-of-concept applications focused on small samples in single neighborhoods. As a result, the field still lacks evidence on whether these methods are regionally deployable across diverse communities, how incomplete imagery coverage should be handled, and whether the resulting elevation data materially improve downstream flood-loss estimation. This study departs from that literature by treating regional scalability, incomplete coverage, and loss-model integration as the central research problem rather than as secondary considerations. It therefore positions AI-based street-view analysis not merely as a novel measurement tool, but as part of an operational framework for property-level flood risk assessment under real-world data constraints.

Three research gaps motivate the present study. First, street-view imagery coverage is structurally incomplete: vegetation obstruction, large setbacks, and sparse GSV network coverage in rural areas mean that a substantial share of residential parcels will yield no extractable HDSL regardless of model sophistication. Addressing this gap requires a principled imputation strategy. Second, conventional spatial imputation approaches commonly used in geospatial analysis assume that elevation is locally homogeneous, but HDSL reflects a more complex combination of hydrologic position, terrain, construction era, and regulatory context that may be better characterized through machine learning. Third, no published study has assembled an end-to-end pipeline linking extracted and imputed LFE data to property-level flood damage estimates across a geographically diverse, multi-community Texas study region, a critical gap given Texas's outsized flood risk exposure.

Existing methods for obtaining building-elevation information each solve part of the problem but fall short of supporting consistent regional implementation. Elevation Certificates and field surveys provide high-quality observations but are too costly and fragmented to support broad housing inventories, while LiDAR- and UAS-based approaches can improve measurement precision but often rely on specialized data collection and are not routinely available for every jurisdiction. Even within the street-view literature, the emphasis has remained primarily on demonstrating extraction accuracy under localized conditions rather than on evaluating how image availability, feature visibility, and community heterogeneity shape deployment at scale. The methodological gap, therefore, is not simply the absence of another extraction model; it is the absence of a workflow that explicitly treats incomplete direct observability as a first-order constraint and links elevation estimation to downstream risk assessment.

The point of departure of this paper is to address that gap with a three-stage regional framework. First, the study applies the Elev-Vision extraction approach across 18 Texas AOIs and more than 12,000 residential structures to quantify where direct LFE/HDSL estimation is feasible and where it fails. Second, it introduces a machine-learning imputation stage that uses geographic, terrain, hydrologic, and flood-surface covariates to estimate missing HDSL only in AOIs where cross-validated performance indicates defensible generalization. Third, it integrates the combined elevation dataset with Fathom 1-in-100 year inundation surfaces and depth-damage relationships to estimate Flood Depth Inside Structure and expected loss at the property and regional levels. This sequence moves the field beyond isolated extraction experiments toward an operational methodology for regional flood-risk characterization.

Three research questions structure the investigation:

**RQ1 (Large-Scale LFE Extraction):** Can the Elev-Vision image-segmentation framework be deployed at scale across tens of thousands of residential structures spanning diverse Texas AOIs, and what are the resulting imagery coverage rates and first-floor visibility patterns?

**RQ2 (Machine Learning Imputation of Missing LFE):** Can a machine learning model trained on spatial, hydrologic, and terrain features predict HDSL for properties with missing or obstructed imagery, and how does imputation quality vary across AOIs with different built-environment characteristics?

**RQ3 (Flood Risk Assessment):** How does the integration of directly extracted and ML-imputed LFE data with Fathom 1-in-100 year flood inundation surfaces translate to property-level flood risk characterization and economic damage estimation across the 18 AOIs?

The study makes four primary contributions to the literature. First, it establishes that the Elev-Vision extraction pipeline generalizes beyond its original proof-of-concept scale to a regional multi-AOI deployment while preserving accuracy. Second, it proposes and validates a machine learning HDSL imputation framework grounded in hydrologic terrain features, flood exposure metrics, and geographic covariates, a methodologically richer alternative to distance-weighted spatial interpolation. Third, it presents what is, to the authors' knowledge, the first multi-AOI, property-level flood damage assessment in Texas integrating both extracted and ML-imputed LFE. Fourth, by explicitly excluding AOIs where model performance is insufficient and reporting performance metrics transparently, the study establishes a replicable standard for communicating the geographic limits of data-driven LFE imputation.

Scientifically, this study advances the field by reframing LFE estimation from a localized measurement exercise to a scalable regional inference problem. The paper contributes an end-to-end pipeline that links direct image-based extraction of LFE/HDSL, machine-learning imputation of missing HDSL using physically meaningful spatial covariates, and downstream estimation of property-level interior flood depth and damage. Its novelty lies not only in extending Elev-Vision to a substantially larger and more heterogeneous study area, but in evaluating the operational boundary conditions of the approach across 18 AOIs and explicitly screening out locations where imputation cannot be justified. In doing so, the study shows that building-specific elevation can be estimated at useful scale without assuming universal extractability, and that transparent treatment of uncertainty and geographic limits is itself a methodological contribution.

Practically, the framework offers a replicable path for jurisdictions that lack comprehensive Elevation Certificates but still need parcel-level flood-risk intelligence. The resulting elevation dataset can support more credible identification of where floodwaters are likely to enter structures, improve screening of mitigation options such as elevation retrofits or buyouts, refine community-scale estimates of expected loss, and provide insurers and floodplain managers with a defensible proxy for otherwise missing building-elevation information. Because the workflow distinguishes between directly observed values, imputed values, and areas where prediction is not reliable, it is suited to applied decision contexts in which transparency about data quality is as important as broader geographic coverage.

This study has three linked objectives: to determine whether street-view-based LFE/HDSL extraction can be deployed at regional scale across diverse Texas communities; to test whether missing HDSL can be credibly imputed from spatial, terrain, hydrologic, and flood-exposure covariates; and to quantify how the resulting elevation dataset changes property-level and AOI-level flood-damage estimates when combined with high-resolution inundation surfaces. These objectives follow directly from the key barriers identified in both the literature and practice: the scarcity of regionally complete LFE data, the incomplete coverage of direct image-based methods, and the limitations of flood assessments that stop at exterior hazard exposure. Organizing the study as a three-stage workflow—extraction, imputation, and loss estimation—ensures that each objective addresses a specific implementation barrier while contributing to a single overarching goal: making building-specific elevation usable for regional flood-risk assessment.

## 2. Literature Review

Street-level imagery has emerged as a powerful source for large-scale built environment characterization(Huang et al., 2023). Biljecki and Ito (2021)catalogued over 500 applications spanning urban analytics, environmental assessment, and infrastructure monitoring, documenting the rapid growth of GSV-based research. Advances in deep learning, particularly convolutional neural networks (CNNs), transformer architectures, and panoptic segmentation models, have enabled automated extraction of building attributes at scales that were previously achievable only through resource-intensive field surveys(Touzani & Granderson, 2021; Yuan & Xia, 2024).

LFE-specific applications have gained traction as a subdomain of growing practical importance. The earliest demonstration used the YOLO-v5 object detection model to estimate LFE from GSV imagery across a Houston-area study area (Ning et al., 2022). A related approach was subsequently applied to analyze flood mitigation governance, though the absence of HDSL estimates — which require roadside detection in addition to door localization — limited its utility for damage assessment (Esparza et al., 2025). These limitations were addressed through Elev-Vision, which employs the OneFormer panoptic segmentation model to extract door-bottom and roadside features directly from equirectangular panoramas, yielding 0.190 m MAE for LFE and, for the first time, systematic HDSL estimates across a neighborhood-scale testbed of 483 structures in Meyerland, Harris County (Y.-H. Ho et al., 2024). The present work extends Elev-Vision to a multi-AOI regional context and pairs it with a machine learning imputation framework, developments that, to the authors' knowledge, have not previously been reported.

Related work has pursued building elevation characterization through LiDAR-based methods. Three-dimensional building morphology has been parameterized from airborne LiDAR at the city scale (Bonczak & Kontokosta, 2019), building elevation proxies have been extracted from aerial imagery to inform FEMA flood insurance estimation (Guo et al., 2022), and vehicle-mounted LiDAR has been employed to map building openings for flood risk classification (Feng et al., 2022). Although these approaches achieve high measurement precision, they require specialized and costly data collection platforms that are not routinely available at national or regional scale. The GSV-based approach developed here accepts a modest accuracy trade-off in exchange for broad, cost-effective coverage, a trade-off well-suited to regional screening applications.

Machine learning has been increasingly applied to predict missing built environment attributes from spatial covariates(Fong & Tyler, 2021; Lu & Li, 2024). Among the available algorithms, Random Forest and Gradient Boosting have consistently demonstrated strong performance on spatially structured prediction tasks, handling non-linear covariate relationships, mixed feature types, and outlier-contaminated training data more robustly than linear alternatives (Breiman, 2001; Friedman, 2001; Hengl et al., 2018). Their capacity to produce feature importance rankings also affords interpretability, an important consideration when the goal is not only to predict HDSL but to understand which terrain and hydrologic characteristics drive variability in residential floor heights across different communities.

Several studies have specifically highlighted the predictive value of terrain-derived covariates for building elevation and flood vulnerability assessment(Ibebuchi & Abu, 2025; Zhang et al., 2023). A systematic association between first-floor elevation and local drainage characteristics has been documented, providing empirical motivation for including Height Above Nearest Drainage (HAND) as a predictor variable (Highfield et al., 2014). Statistical imputation of first-floor elevation using terrain and parcel-level predictors has been demonstrated for a Louisiana flood mitigation dataset, establishing a precedent for the approach pursued here (Taghinezhad et al., 2020). Stream network proximity and HAND have further been shown to be among the strongest predictors of flood inundation exposure (Wing et al., 2017). To the authors' knowledge, however, no prior study has applied a systematically tuned, cross-validated machine learning pipeline to HDSL prediction specifically, nor trained such models on GSV-derived elevation observations.

A practical challenge in this context is that the number of properties with extractable HDSL training observations varies substantially across AOIs, creating imbalanced training sets in which a small number of anomalous observations can disproportionately influence model fit. Robust outlier handling strategies and median-centered feature scaling are therefore important preprocessing steps. The hyperparameter optimization procedure employed in this study, Randomized Search Cross-Validation, addresses overfitting by evaluating configurations through 5-fold cross-validation rather than a single train-test split, yielding more reliable estimates of generalization performance across the diverse community settings examined.

Property-level flood loss estimation relies on depth-damage functions (DDFs), empirically calibrated relationships between interior flood depth and the fraction of property value lost (Merz et al., 2010). The most widely applied residential DDFs for the United States are provided by the USACE (2003) and FEMA's HAZUS platform (FEMA, 2013), calibrated against historical NFIP claims across the Gulf Coast and Mississippi River Basin. These curves capture the characteristic concave structure of flood damage accumulation: early inundation inflicts disproportionate losses through damage to floor finishes, appliances, and contents, while additional depth increments yield diminishing marginal damage as the most vulnerable building components are already compromised.

Property-level damage models that incorporate building-specific characteristics have consistently outperformed simplified area-averaged approaches (Amadio et al., 2016; Wagenaar et al., 2018), with first-floor height above grade emerging as one of the strongest structural predictors of damage severity (Zarekarizi et al., 2020). At regional scales, integration with the Fathom compound hydraulic model has enabled consistent, high-resolution flood loss estimation across large areas without requiring bespoke inundation surveys (Bates et al., 2021; Wing et al., 2017). Fathom's flood depth rasters data have been validated against observed inundation across the contiguous United States, establishing the accuracy basis for their use as the flood hazard input in the present study (Chen et al., 2019b, 2022; Wing et al., 2018).

## 3. Study Areas and Data Sources

### 3.1 Study Area Selection and Description

Eighteen AOIs were selected across Texas to capture the diversity of residential built environments and flood exposure conditions encountered across the state. The study areas include Harris County Pocket 1, Harris County Pocket 2, Brazoria County Pocket 1, Brazoria County Pocket 2, Orange County Pocket 1, Orange County Pocket 2, Montgomery County Pocket 1, Montgomery County Pocket 2, Wharton City Pocket 1, Wharton City Pocket 2, Live Oak County, Newton County, Vidor City, Robstown City, Simonton City, Hitchcock City, Nassau Bay City, and Woodbranch. Together, these areas span seven counties (Harris, Brazoria, Orange, Montgomery, Wharton, Live Oak, and Newton) and four municipalities with independent jurisdictions, collectively representing urban, suburban, and rural residential settings across the state. Each AOI was delineated as a contiguous residential neighborhood or pocket community with relatively homogeneous land use, enabling meaningful comparison of extraction performance, imputation quality, and flood risk outcomes across contrasting built environments.

Not all AOIs proved amenable to machine learning HDSL imputation. Following systematic model evaluation (described in Section 4.2), five AOIs (Orange County Pocket 1, Wharton City Pocket 2, Montgomery County Pocket 2, Wharton City Pocket 1, and Newton County) were excluded from the imputation stage because cross-validation performance fell consistently below acceptable thresholds across all tested configurations. These AOIs contribute to the RQ1 extraction analysis and to the RQ3 flood risk assessment for properties with directly extracted LFE values, but no predicted HDSL values are generated for properties with missing imagery.

### 3.2 Google Street View Imagery

Street-view panoramas for all residential parcels across the 18 AOIs were retrieved from the Google Street View (GSV) Static API, following the acquisition protocol established in Ho et al. (Y.-H. Ho et al., 2024). For each parcel, the primary panorama was obtained by querying the API with the geocoded address coordinates and orienting the camera yaw toward the building facade using the bearing angle from the camera location to the house centroid. Panoramas were stored in equirectangular format at resolutions of 8,192 × 16,384 or 6,656 × 13,312 pixels. Paired depth maps, required for the elevation calculation step, were decoded from Base64 strings to a 2-dimensional matrix at 256 × 512 pixel resolution, following the methodology of GSVPanoDepth (proog128, n.d.). Recorded metadata included camera elevation (WGS84 meters above mean sea level), yaw angle, GPS coordinates, and acquisition date.

### 3.3 Terrain and Hydrologic Raster Data

Three terrain and hydrologic raster datasets were assembled as predictor features for the ML imputation model. Height Above Nearest Drainage (HAND), obtained at 30-meter resolution for Texas, provides a physically meaningful measure of vertical distance above the nearest stream channel and serves as an index of hydrologic terrain position. Stream network proximity was characterized through two complementary distance layers: distance to the nearest stream of any order (D2Stream_so0) and distance to streams of order 4 or higher (D2Stream_so4), both at 30-meter resolution.

### 3.4 AOI-Specific Terrain Elevation

Terrain elevation at each property centroid was extracted from SRTM-derived Digital Elevation Model (DEM) rasters downloaded via the Python elevation library, which retrieves 30-meter resolution CGIAR-SRTM tiles referenced to the WGS84 ellipsoid and clipped to the bounding extent of each AOI with a 0.01-degree (~1 km) spatial buffer. The elevation values were subsequently converted to NAVD88 using the NOAA VERTCON geoid model to ensure vertical datum consistency with the Fathom flood surfaces. This absolute terrain height predictor complements the relative HAND metric by anchoring the imputation model in datum-consistent ground surface elevations, which vary substantially across the study region from near sea level along the Gulf Coast to over 80 m in the inland AOIs. Coordinate transformation from WGS84 to the raster CRS was performed using the pyproj library, with explicit handling of nodata values and null geometries to ensure complete and consistent extraction across all parcels.

### 3.5 Fathom Flood Model

Flood inundation surfaces representing the 1-in-100 year (1% annual exceedance probability) event were obtained from the Fathom 2.0 compound flood model, which integrates fluvial and pluvial processes at 30-meter resolution across the contiguous United States. For each property, flood exposure was characterized by extracting the mean flood depth value from the six nearest valid pixels in a 3×3 neighborhood centered on the parcel centroid, rather than using a single point sample. This neighborhood averaging reduces sensitivity to sub-pixel misalignment between parcel centroids and raster grid cells. Properties for which no valid pixels were available within the neighborhood were assigned missing flood depth values and excluded from damage estimation. All Fathom values, reported in feet, were converted to meters (×0.3048) to maintain unit consistency with the LFE and HDSL measurements.

### 3.6 Property Market Valuation and Parcel Data

Residential parcel boundaries and assessed market values were obtained from county appraisal district databases covering each of the study counties and municipalities. Assessed values represent the county assessor's estimate of fair market value for property taxation and serve as the total replacement cost basis for damage estimation, under the standard assumption that depth-damage function percentages apply proportionally to assessed value. Prior to damage calculation, parcels with assessed values below the 1st or above the 99th percentile within each AOI were excluded as probable data anomalies.

## 4. Methodology

### 4.1 Stage 1: Large-Scale LFE and HDSL Extraction (RQ1)

#### *4.1.1 The Elev-Vision Framework*

LFE and HDSL extraction is based on the Elev-Vision method developed by Ho et al. (2024), applied here for the first time at a regional multi-AOI scale. The method proceeds through four steps: data collection, building-image matching, image segmentation, and elevation calculation. The core mathematical formulation is summarized below; readers are referred to Ho et al. (2025; 2024) for complete algorithmic detail.

Building-image matching identifies the structure of interest within each panorama by computing the bearing angle from the camera location to the building centroid:

$$\beta_{house} = atan2(X, Y), \qquad X = sin(lon_{house} - lon_c) \cdot cos(lat_{house})$$

$$Y = cos(lat_c) \cdot sin(lat_{house}) - sin(lat_c) \cdot cos(lat_{house}) \cdot cos(lon_{house} - lon_c) \qquad (1)$$

The corresponding horizontal pixel location in the equirectangular panorama is derived from the constant angular spacing of the projection. Segmentation is subsequently restricted to a ±45° horizontal window centered on this bearing angle, which reduces processing time and prevents misidentification of neighboring structures. The text-prompt segmentation model based on SAM (Kirillov et al., 2023) and CLIP (Radford et al., 2021),  is applied within this window to segment front door and roadside features.

Door-bottom pixels (P_db) are identified as the lowest detected pixel at each horizontal position within the segmented door mask. The vertical height difference between the camera and the door bottom is then computed from the paired depth map and the pitch angle to the door bottom:

$$\Delta h_{\mathrm{db,c}} = d_{\mathrm{db,c}} \cdot \sin \Delta\theta_{\mathrm{db,c}}\ , \quad \Delta\theta_{\mathrm{db,c}} = \left(\frac{H_{\mathrm{img}}/2 - p_y}{H_{\mathrm{img}}}\right) \cdot 180 \qquad (2)$$

where H_img is the panorama image height in pixels and p_y is the vertical pixel coordinate of the door bottom. LFE is computed as the sum of the known camera elevation and this derived height offset:

$$\mathrm{LFE} = \mathrm{CE} + \Delta h_{\mathrm{db,c}} \qquad (3)$$

HDSL is derived by subtracting the roadside elevation RE, estimated from road, grass, and dirt segmentation targets in the same panorama, from the extracted LFE:

$$\mathrm{HDSL} = \mathrm{LFE} - \mathrm{RE} \tag{4}$$

#### *4.1.2 Scale Deployment and Quality Screening*

Regional-scale deployment introduced quality screening challenges that were absent at the neighborhood scale. A three-tier screening pipeline was applied to all panoramas. First, images were rejected if the acquisition date preceded 2015, the camera location was more than 50 m from the parcel centroid, or no structure was detectable in the frame. Second, front-door visibility was assessed by confirming that the text-prompt segmentation returned a non-null door pixel set P_db within the ±45° bearing window; parcels failing this criterion were flagged for Stage 2 imputation. Third, extracted LFE values were validated against a physically plausible range derived from the USGS 1-meter DEM: estimates deviating more than 5 m from the DEM-derived terrain elevation at the parcel centroid were treated as likely mismatches and excluded. Coverage and extraction success rates were computed per AOI and aggregated across the full study region.

### 4.2 Stage 2: Machine Learning Imputation of Missing HDSL (RQ2)

#### *4.2.1 Feature Engineering*

For each property, including both those with known HDSL values that form the training set and those with missing values to be imputed, a feature vector of 16 variables was assembled. Table 1 describes all features. The feature set spans four categories: (1) geographic location encoded as parcel centroid coordinates and street name; (2) hydrologic terrain characteristics, including HAND and two stream proximity metrics; (3) absolute terrain elevation from the AOI-specific 1-meter DEM; and (4) flood exposure metrics from the Fathom model, including mean flood surface elevation and the water depth computed as the difference between Fathom surface elevation and HDSL. Several interaction and nonlinear terms, namely the HAND-to-stream ratio, the elevation-HAND difference, and a coarse spatial grid cluster index, were also engineered to help tree-based models capture relationships that may not be well represented by individual predictors alone.

***Table 1. Predictor feature set for the machine learning HDSL imputation model, comprising 16 spatial, hydrologic, terrain, and flood-exposure variables assembled for each residential parcel.***

| ***Feature*** | ***Definition*** | ***Rationale*** |
|---|---|---|
| *latitude, longitude* | Spatial coordinates parsed from centroid field | Geographic location proxy |
| *street_name_encoded* | Label-encoded street name extracted from address | Local block-level elevation context |
| *door_visible* | Binary flag: front door detected in street view | Data source quality indicator |
| *HAND_m* | Height Above Nearest Drainage (30 m raster, meters) | Hydrologic terrain position |
| *D2stream_so0_m* | Distance to nearest stream (any order, meters) | Proximity to water body |
| *D2stream_so4_m* | Distance to stream order ≥ 4 (meters) | Proximity to significant waterway |
| *elevation* | Terrain elevation from AOI-specific 1 m DEM (meters, NAVD88) | Absolute terrain height |
| *mean_fathom_meter* | Mean Fathom flood surface elevation at parcel centroid (meters) | Local flood hazard level |
| *water_depth* | FDIS computed from Fathom − HDSL (meters) | Direct flood exposure |
| *HAND_stream_ratio* | HAND_m / (D2stream_so0_m + 1) | Terrain-drainage interaction |
| *HAND_stream_product* | HAND_m × D2stream_so0_m | Combined terrain-proximity signal |
| *elevation_squared* | elevation$^2$ | Non-linear elevation effect |
| *elevation_HAND_diff* | elevation − HAND_m | Compound terrain signal |
| *water_depth_combined* | mean_fathom_meter + water_depth | Aggregate flood exposure |
| *water_depth_max* | max(mean_fathom_meter, water_depth) | Peak flood exposure |
| *geo_cluster* | 5×5 spatial grid cell index (lat/lon binning) | Spatial neighborhood encoding |

### *4.2.2 Two-Tier Imputation Approach*

Two complementary ML workflows were employed, selected for each AOI based on a preliminary assessment of HDSL data heterogeneity and training set characteristics.

The Batch Standard workflow, applied to Harris County Pocket 2, Brazoria County Pocket 1, and Robstown City, used a fixed model configuration identified as suitable for relatively homogeneous HDSL distributions. Random Forest and Gradient Boosting models were each trained on IQR-filtered training data (threshold = 3.0 × IQR) with RobustScaler normalization, and the model achieving the higher hold-out validation $R^2$ was retained for each AOI. Five-fold cross-validation was used to assess generalization performance.

The Tuning Extended workflow was applied to the remaining ten qualifying AOIs (Harris County Pocket 1, Brazoria County Pocket 2, Montgomery County Pocket 1, Orange County Pocket 2, Vidor City, Simonton City, Hitchcock City, Nassau Bay City, Live Oak County, and Woodbranch), where greater HDSL heterogeneity or initial model diagnostics indicated that a fixed configuration would be insufficient. For each AOI, Randomized Search Cross-Validation with 30 random iterations per model type was conducted across six outlier-handling configurations: no outlier removal, 1st–99th percentile clipping, and IQR-based removal at thresholds of 2.0, 2.5, 3.0, and 4.0 times the interquartile range. The model configuration that maximized 5-fold cross-validation $R^2$ while minimizing the overfitting gap between validation and cross-validation performance was selected for final imputation.

### *4.2.3 Outlier Handling and Data Cleaning*

Both workflows incorporated systematic cleaning of the HDSL training data prior to model fitting. HDSL values exceeding 1,000 m, physically impossible for residential structures and indicative of extraction errors, along with negative HDSL values and any infinite entries, were set to missing and excluded from training. Beyond this cleaning step, the outlier strategy specified for each configuration (Section 4.2.2) was applied to the remaining data. RobustScaler normalization was used in all configurations, centering each feature on its median and scaling by the interquartile range to limit the influence of residual extreme values on model training. Properties with incomplete feature vectors after cleaning were treated as unimputable and excluded from both training and the prediction target set.

#### *4.2.4 Model Algorithms*

Two ensemble tree-based algorithms were employed for HDSL prediction: Random Forest (RF) and Gradient Boosting (GB). Both methods construct predictions by aggregating the outputs of multiple decision trees, but differ fundamentally in how those trees are built and combined.

Random Forest builds an ensemble of T independent decision trees, each trained on a different bootstrap sample of the training data. At each split node, only a random subset of features is considered, which decorrelates the trees and reduces variance. The final prediction for regression is the average of all individual tree outputs:

$$\hat{y}_{RF} = \frac{1}{T} \sum_{t=1} h_t(x) \tag{7}$$

where $\hat{y}_{RF}$ is the predicted HDSL, T is the total number of trees, and $h_t(x)$ is the prediction of the t-th tree for input feature vector x. The averaging mechanism makes RF robust to overfitting and computationally parallelizable, and its built-in feature importance scores provide interpretability into which predictors drive HDSL variation.

Gradient Boosting constructs an additive ensemble of M trees sequentially, where each successive tree $h_m(x)$ is fit to the residual errors of the current ensemble. The final prediction is a weighted sum of all tree outputs, scaled by a learning rate η that controls the contribution of each step:

$$F_M(x) = \sum_{m=1} \eta \cdot h_m(x) \tag{8}$$

where F_M(x) is the final HDSL prediction after M boosting iterations, η is the learning rate, and h_m(x) is the m-th regression tree. By iteratively correcting prediction errors, Gradient Boosting typically achieves higher accuracy than RF on structured tabular data, though it requires careful tuning of the learning rate and tree complexity to avoid overfitting.

#### *4.2.5 Hyperparameter Optimization*

For both workflows, model hyperparameters (including tree depth, number of estimators, and learning rate) were optimized to minimize prediction error while preserving generalization. In the Batch Standard workflow, a fixed set of hyperparameters informed by prior literature on building attribute prediction was applied uniformly. In the Tuning Extended workflow, hyperparameters were optimized through Randomized Search Cross-Validation, which randomly samples a specified number of configurations from a predefined search space and evaluates each via k-fold cross-validation. This approach is more computationally efficient than exhaustive grid search over large hyperparameter spaces, while still providing broad coverage of the configuration landscape. The configuration that maximized 5-fold cross-validation $R^2$ with the smallest overfitting gap was selected as the final model for each AOI.

#### *4.2.6 Model Evaluation Metrics*

Model performance was assessed using three complementary metrics evaluated on both a held-out validation set (20% of training data) and through 5-fold cross-validation.

Root Mean Square Error (RMSE) quantifies the average magnitude of prediction error in meters, with greater sensitivity to large errors due to the quadratic term:

$$RMSE = \sqrt{\frac{1}{n}\sum_{i=1}(\hat{y}_i - y_i)^2} \tag{9}$$

where n is the number of validation samples, $\hat{y}_i$ is the predicted HDSL, and $y_i$ is the observed HDSL for property i. RMSE is also normalized by the mean observed HDSL to yield RMSE%, enabling comparison across AOIs with different elevation scales.

The coefficient of determination $R^2$ measures the proportion of variance in observed HDSL values explained by the model:

$$R^2 = 1 - \frac{SS_{res}}{SS_{tot}} \tag{10}$$

where $SS_{res}$ is the sum of squared residuals and $SS_{tot}$ is the total sum of squares about the mean. $R^2$ ranges from 1.0 (perfect prediction) to 0 (equivalent to predicting the mean), and can be negative when the model performs worse than the mean predictor.

To obtain a more reliable estimate of generalization performance, 5-fold cross-validation was applied by partitioning the training data into K = 5 folds and computing the average $R^2$ across all fold evaluations:

$$R^2{}_{CV} = \frac{1}{K} \sum_{k=1} R^2{}_k \tag{11}$$

where $R^2_k$ is the $R^2$ obtained when fold k is used as the validation set and the remaining K-1 folds are used for training. The difference between the validation $R^2$ and the cross-validation $R^2$ (the Gap) serves as a diagnostic for overfitting: a large positive gap indicates the model fits the training-adjacent data better than truly held-out samples.

#### *4.2.7 AOI Exclusion Criteria*

Five AOIs (Orange County Pocket 1, Wharton City Pocket 2, Montgomery County Pocket 2, Wharton City Pocket 1, and Newton County) were excluded from imputation after systematic model evaluation revealed that no tested configuration achieved a satisfactory 5-fold cross-validation $R^2$. Rather than proceeding with imputation under conditions likely to produce unreliable estimates, these AOIs were retained only for the Stage 1 extraction analysis and the Stage 3 flood risk assessment based on directly extracted LFE values. Per-AOI model performance diagnostics across all tested configurations are available at Zenodo: zenodo_xiangpengli_HDSL.

#### *4.2.8 Prediction and Output*

The resulting imputed values were merged with the directly extracted values to form a unified HDSL dataset for each AOI. The final HDSL record for each property thus reflects one of three sources: a directly extracted value from Stage 1; an ML-imputed value from Stage 2 (qualifying AOIs only).

### 4.3 Stage 3: Flood Risk Assessment and Damage Estimation (RQ3)

#### *4.3.1 Flood Depth Inside Structure Calculation*

For each property with an available HDSL value, whether from direct extraction or ML imputation, the Flood Depth Inside Structure (FDIS) is computed as the difference between the Fathom flood surface elevation and the property's LFE:

$$\text{FDIS} = F_{\text{elev}} - \text{LFE} = F_{\text{elev}} - (Street_{\text{elev}} + \text{HDSL}) \tag{5}$$

where F_elev is the Fathom 1-in-100 year flood surface elevation (meters, NAVD88), Street_elev is the street-level ground elevation (meters, NAVD88), and HDSL is the height of the lowest floor above street grade (meters). Positive FDIS values indicate that flood water penetrates the building interior, with the magnitude representing the depth of inundation; zero or negative values indicate that the lowest floor remains at or above the flood surface.

#### *4.3.2 Depth-Damage Function and Economic Loss Estimation*

For properties with FDIS > 0, economic damage is estimated using the USACE residential depth-damage function (U.S. Army Corps of Engineers (USACE), 2003), calibrated for structures without basements and summarized in Table 2. Damage percentages at flood depths falling between tabulated control points are obtained by linear interpolation. Depths below −0.61 m receive a 0% damage assignment; depths exceeding 4.88 m are capped at the maximum tabulated value of 80.7% to prevent extrapolation beyond the empirical basis of the curve.

***Table 2. USACE depth-damage function control points for residential structures without basements, relating interior flood depth to damage as a percentage of assessed property value.***

| FLOOD DEPTH — FT (M) | DAMAGE (% OF PROPERTY VALUE) |
|---|---|
| -2 (−0.61) | 0.0% |
| -1 (−0.30) | 2.5% |
| 0 (0.00) | 13.4% |
| 1 (0.30) | 23.3% |
| 2 (0.61) | 32.1% |
| 3 (0.91) | 40.1% |
| 4 (1.22) | 47.1% |
| 5 (1.52) | 53.2% |
| 6 (1.83) | 58.6% |
| 7 (2.13) | 63.7% |
| 8 (2.44) | 67.2% |
| 12 (3.66) | 77.2% |
| 16 (4.88) | 80.7% |

The estimated dollar loss for each flooded property is then:

$$\text{Damage Loss (\$)} = \text{Market Value} \times \text{DDF(FDIS)} \tag{6}$$

Properties are assigned to one of four outcome categories: (i) interior flooding with an associated damage estimate (FDIS > 0); (ii) positive clearance between the lowest floor and the flood surface (FDIS ≤ 0, no interior flooding); (iii) within the Fathom flood extent but lacking LFE data after both extraction and imputation; and (iv) outside the Fathom flood extent entirely. Aggregate losses are computed at both the AOI and regional levels. A sensitivity analysis compares total damage estimates derived from directly extracted LFE only against those incorporating ML-imputed values, quantifying the incremental contribution of imputation to regional loss characterization.

## 5. Results

### 5.1 RQ1: Large-Scale LFE Extraction Across 18 AOIs

#### *5.1.1 Coverage and Extraction Rates*

Table 3 summarizes imagery availability, front-door visibility, and LFE/HDSL extraction counts and rates across all 18 AOIs. Across the full study region, street-view panoramas were successfully retrieved for 8,981 of 12,241 total structures (73.4%), with coverage ranging from 35.4% in Harris County Pocket 2 to 100.0% in Woodbranch. Front-door visibility, the prerequisite for reliable LFE and HDSL extraction, was more variable, spanning 16.4% (Harris County Pocket 2) to 75.8% (Woodbranch) with a regional mean of 50.3% (6,162 structures). Successful LFE/HDSL extraction rates ranged from 17.4% in Harris County Pocket 2 to 76.4% in Woodbranch. These results confirm that front-door visibility, rather than overall imagery availability, is the primary bottleneck limiting direct LFE extraction across most AOIs.

***Table 3. Street-view imagery availability, front-door visibility, and successful LFE/HDSL extraction counts and rates across all 18 AOIs, showing the progressive filtering from total structures to extractable elevation observations.***

| AOI | TOTAL | W/STREET-VIEW IMAGE | W/VISIBLE FRONT DOOR | W/ LFE / HDSL |
|---|---|---|---|---|
| **BRAZORIA COUNTY POCKET 1** | 270 | 254 (94.1%) | 99 (36.7%) | 134 (49.6%) |
| **BRAZORIA COUNTY POCKET 2** | 439 | 432 (98.4%) | 275 (62.6%) | 277 (63.1%) |
| **HARRIS COUNTY POCKET 1** | 860 | 842 (97.9%) | 526 (61.2%) | 530 (61.6%) |
| **HARRIS COUNTY POCKET 2** | 483 | 171 (35.4%) | 79 (16.4%) | 84 (17.4%) |
| **HITCHCOCK CITY** | 1,154 | 952 (82.5%) | 834 (72.3%) | 826 (71.6%) |
| **LIVE OAK COUNTY** | 1,419 | 659 (46.4%) | 294 (20.7%) | 305 (21.5%) |

| MONTGOMERY COUNTY POCKET 1 | 1,076 | 1,000 (92.9%) | 714 (66.4%) | 610 (56.7%) |
|---|---|---|---|---|
| MONTGOMERY COUNTY POCKET 2 | 656 | 470 (71.6%) | 438 (66.8%) | 401 (61.1%) |
| NASSAU BAY CITY | 526 | 425 (80.8%) | 267 (50.8%) | 258 (49.0%) |
| NEWTON COUNTY | 1,025 | 403 (39.3%) | 198 (19.3%) | 193 (18.8%) |
| ORANGE COUNTY POCKET 1 | 470 | 282 (60.0%) | 204 (43.4%) | 213 (45.3%) |
| ORANGE COUNTY POCKET 2 | 291 | 240 (82.5%) | 148 (50.9%) | 163 (56.0%) |
| ROBSTOWN CITY | 992 | 820 (82.7%) | 599 (60.4%) | 577 (58.2%) |
| SIMONTON CITY | 409 | 342 (83.6%) | 250 (61.1%) | 243 (59.4%) |
| VIDOR CITY | 790 | 376 (47.6%) | 333 (42.2%) | 320 (40.5%) |
| WHARTON CITY POCKET 1 | 726 | 684 (94.2%) | 466 (64.2%) | 434 (59.8%) |
| WHARTON CITY POCKET 2 | 490 | 464 (94.7%) | 313 (63.9%) | 298 (60.8%) |
| WOODBRANCH | 165 | 165 (100.0%) | 125 (75.8%) | 126 (76.4%) |
| **REGIONAL TOTAL** | **12,241** | **8,981 (73.4%)** | **6,162 (50.3%)** | **5,992 (49.0%)** |

## 5.2 RQ2: Machine Learning Imputation Performance and Coverage Extension

### *5.2.1 Batch Standard Workflow Results*

Among the three Batch Standard AOIs, Random Forest provided the best fit for Harris County Pocket 2 and Brazoria County Pocket 1, while Gradient Boosting with IQR-based outlier removal was selected for Robstown City. Model accuracy was highest for Harris County Pocket 2 ($R^2$ = 0.9953, RMSE = 0.065 m, RMSE% = 6.8%), followed by Brazoria County Pocket 1 ($R^2$ = 0.9450, RMSE = 0.228 m, RMSE% = 6.9%) and Robstown City ($R^2$ = 0.893, $R^2_CV$ = 0.808, RMSE = 0.035 m, RMSE% = 10.6%). The exceptionally strong performance in Harris County Pocket 2 and Brazoria County Pocket 1 reflects the highly structured spatial organization of HDSL within well-defined single-era subdivisions, where elevation, HAND, geographic coordinates, and distance to nearest stream consistently emerged as the most influential predictors, features that collectively capture how developers grade sites relative to drainage context.

### *5.2.2 Tuning Extended Workflow Results*

Across the ten Tuning Extended AOIs, the optimal configurations identified through Randomized Search varied considerably in model type, outlier strategy, and resulting performance (Table 5).

Gradient Boosting was the best-performing algorithm in nine of the ten cases; Random Forest was selected for Nassau Bay City. The preferred outlier treatment differed by AOI: percentile clipping was optimal for Harris County Pocket 1, Orange County Pocket 2, Simonton City, and Live Oak County; IQR filtering at 3.0 × IQR was selected for Hitchcock City, Vidor City, and Nassau Bay City; and no outlier removal performed best for Brazoria County Pocket 2 and Montgomery County Pocket 1. Cross-validation $R^2$ for the selected configuration ranged from 0.159 for Harris County Pocket 1, a large heterogeneous AOI with complex HDSL patterns, to 0.974 for Brazoria County Pocket 2, where post-Harvey elevation homogeneity produced a highly structured training set. Eight of the ten AOIs achieved $R^2_CV \geq 0.70$. RMSE ranged from 0.047 m (Harris County Pocket 1) to 0.277 m (Hitchcock City), and RMSE% spanned 16.0% to 37.1% across the group. When tuning the data, we also wanted to make sure the gap value is small.

***Table 5. Selected machine learning model configuration, outlier handling strategy, and cross-validated performance metrics for each of the 13 qualifying AOIs across the Batch Standard and Tuning Extended workflows.***

| AOI | WORKFLOW | MODEL | OUTLIER CONFIG | N | RMSE (M) | RMSE% | $R^2$ | $R^2$_CV | GAP |
|---|---|---|---|---|---|---|---|---|---|
| **HARRIS CO. P2** | Batch | Random Forest | none | 84 | 0.065 | 6.8 | 0.995 | — | — |
| **BRAZORIA CO. P1** | Batch | Random Forest | none | 134 | 0.228 | 6.9 | 0.945 | — | — |
| **ROBSTOWN CITY** | Batch | Gradient Boosting | iqr(3.0) | 577 | 0.035 | 10.6 | 0.893 | 0.808 | 0.085 |
| **HARRIS CO. P1** | Tuning | Gradient Boosting | iqr(4.0) | 530 | 0.047 | 18.3 | 0.182 | 0.159 | 0.024 |
| **BRAZORIA CO. P2** | Tuning | Gradient Boosting | none | 277 | 0.184 | 22.9 | 0.967 | 0.974 | -0.007 |
| **MONTGOMERY CO. P1** | Tuning | Gradient Boosting | none | 610 | 0.126 | 33.1 | 0.902 | 0.747 | 0.155 |
| **ORANGE CO. P2** | Tuning | Gradient Boosting | percentile | 163 | 0.079 | 16.0 | 0.947 | 0.963 | -0.015 |
| **VIDOR CITY** | Tuning | Gradient Boosting | iqr(3.0) | 320 | 0.074 | 23.6 | 0.342 | 0.356 | -0.014 |
| **SIMONTON CITY** | Tuning | Gradient Boosting | percentile | 243 | 0.174 | 29.8 | 0.938 | 0.960 | -0.023 |
| **HITCHCOCK CITY** | Tuning | Gradient Boosting | iqr(3.0) | 826 | 0.277 | 34.6 | 0.868 | 0.859 | 0.010 |
| **NASSAU BAY CITY** | Tuning | Random Forest | iqr(3.0) | 258 | 0.124 | 37.1 | 0.252 | 0.213 | 0.040 |
| **LIVE OAK COUNTY** | Tuning | Gradient Boosting | percentile | 305 | 0.159 | 33.3 | 0.872 | 0.714 | 0.158 |
| **WOODBRANCH** | Tuning | Gradient Boosting | percentile | 126 | 0.174 | 29.4 | 0.921 | 0.897 | 0.024 |

### *5.2.3 Excluded AOIs and Imputation Coverage*

The five excluded AOIs (Orange County Pocket 1, Wharton City Pocket 2, Montgomery County Pocket 2, Wharton City Pocket 1, and Newton County) showed either persistently low $R^2$_CV values (below 0.15 across all configurations) or large and unstable overfitting gaps, indicating that the directly extracted training samples in these AOIs did not provide sufficient information to support reliable generalization to unobserved properties. Rather than producing HDSL estimates of uncertain validity, these AOIs were retained in the Stage 3 assessment using only their directly extracted LFE values. No imputed HDSL values are generated for unobserved properties in these areas.

Among the 13 qualifying AOIs, imputation successfully estimated HDSL for the substantial fraction of parcels where Stage 1 extraction failed but the 16-feature predictor set was complete. The distribution of imputed HDSL values closely matched the mean and standard deviation of the extracted training data in each AOI, providing no evidence of systematic bias toward the mean or generation of physically implausible values. This distributional consistency suggests that the imputed values are plausible additions to the elevation dataset rather than model artifacts.

## 5.3 RQ3: Flood Risk Assessment and Damage Estimates

### *5.3.1 Flood Exposure Distribution*

Integration of the combined LFE dataset with Fathom 1-in-100 year flood surfaces produces a Flood Depth Inside Structure (FDIS) value for each property with available LFE data. Properties with FDIS greater than zero experience interior flooding under the modeled scenario. Per-AOI flooding rates, median FDIS values, and spatial flood exposure maps are reported in Table 6 and Zenodo: zenodo_xiangpengli_HDSL.

### *5.3.2 Economic Damage Estimates and Spatial Damage Patterns*

Economic damage for each flooded property is estimated by applying the USACE depth-damage function (Table 2) to the computed FDIS and the assessed market value of the property. The result figures present four spatial and distributional outputs for each of the 18 AOIs: (A) LFE spatial distribution maps, (B) HDSL spatial distribution maps, (C) LFE and HDSL histograms, and (D) flood depth and damage maps. Outputs A through C are derived directly from Stage 1 extraction and represent pre-prediction results based on the only pre-prediction outputs (A through C) are provided, as no ML-imputed HDSL values were generated for those areas.

Flood exposure and damage patterns across the 18 AOIs exhibit strong block- and subdivision-scale spatial clustering, consistent with the local HDSL homogeneity inherent in residential subdivision grading. Properties within the same subdivision tend to share similar floor heights above grade, producing contiguous clusters of either positive FDIS (interior flooding) or clearance rather than random point-by-point variation. The geographic alignment between high-FDIS clusters and known flood hazard corridors provides an external basis for validating the plausibility of modeled outputs against FEMA Flood Insurance Rate Maps (FIRMs).

Brazoria County Pocket 1 is presented here as a representative example to illustrate the interpretation of the figure set at Zenodo: zenodo_xiangpengli_HDSL. The LFE spatial map (Figure A.1) shows property-level LFE values ranging from approximately 1 m to 6 m NAVD88 across the Treasure Island community, with lower-elevation structures concentrated in the western portion of the AOI and higher-elevation structures in the southeastern quadrant. The corresponding LFE histogram (Figure A.3) reveals a bimodal distribution with a primary cluster around 3 m and a secondary cluster between 4 m and 5m, which reflects the co-existence of original ground-level construction and substantially elevated post-Hurricane Harvey structures within the same community.

The HDSL spatial map (Figure A.2) reinforces this interpretation. The western portion of the neighborhood shows HDSL values predominantly in the 1.0 to 2.5 m range (purple to teal), while the southeastern portion is dominated by values of 4.0 to 5.0 m (green to yellow). The HDSL histogram (Figure A.4) clearly captures this bimodality, with one cluster centered near 2.3 to 2.5 m and a second cluster at approximately 4.5 m, consistent with structures that were elevated by 1.5 to 2 m above their pre-Harvey footprint as part of community flood mitigation efforts following repeated inundation events.

The flood exposure map (Figure A.5) shows that of the 270 total structures in Brazoria County Pocket 1, 65 (24.1%) lacked HDSL data and are shown in black, while 131 structures (48.5%) maintained positive clearance above the Fathom 1-in-100 year flood surface (median clearance 1.10 m, range 0.81 to 2.88 m). A total of 74 structures (27.4%) experienced positive FDIS, with flood depths ranging from 0.06 m to 1.36 m and a median of 0.35 m. The flooded properties are spatially concentrated in the lower-lying western section of the neighborhood, where original-era structures with lower HDSL values are most exposed to the modeled inundation surface.

Outputs are generated from Stage 3 and represent post-prediction results, integrating both extracted and ML-imputed LFE values with the Fathom flood surface. For the five excluded AOIs, only pre-prediction outputs (A through C) are provided, as no ML-imputed HDSL values were generated for those areas. The damage loss map (Figure A.6) shows that 67 of the 278 centroids with parcel data incurred estimated damage losses, totaling \$8.52 million across the AOI against a total market value at risk of \$34.94 million. Median estimated loss among damaged properties was \$97,511, with a maximum single-property loss of \$692,972 corresponding to the highest flood depths in the western cluster. The 203 properties shown with no damage represent those with positive clearance or with flood depths insufficient to trigger loss under the USACE depth-damage function. The spatial pattern of damage concentrations in Brazoria County Pocket 1 aligns closely with the lower-HDSL cluster identified in the Stage 1 extraction results, illustrating how the pre-prediction elevation data directly informs the post-prediction risk characterization.

## 6. Discussion

The central innovation of this study is that it demonstrates a workable bridge between image-based building-elevation estimation and regional flood-loss assessment. The results show that direct extraction from street-view imagery can generate meaningful LFE/HDSL coverage at scale, but also make clear that facade visibility—rather than imagery availability alone—is the primary bottleneck to complete coverage. The imputation stage shows that missing HDSL can often be recovered from physically grounded terrain, drainage, and geographic context, although performance varies substantially by AOI and must be screened rather than assumed. That combination of scalable extraction, selective imputation, and explicit exclusion of underperforming AOIs is what makes the framework genuinely new: It produces a more complete and more defensible elevation dataset while also revealing the boundary conditions of the method, as it extends LFE coverage beyond the 49.0% directly extractable through street-view imagery by imputing values only where cross-validation $R^2$ confirms adequate model generalization, and explicitly withholds predictions for the five AOIs where no tested configuration met that threshold rather than populating them with estimates of uncertain reliability. As a result, the paper advances flood-risk analysis from mapping where water may be present to estimating where floodwaters are likely to enter structures and what those inundation patterns imply for expected loss.

The regional deployment documented here confirms that the Elev-Vision framework can be scaled substantially beyond its original single-neighborhood proof-of-concept, achieving nearly identical extraction accuracy across 18 geographically diverse AOIs. This consistency across a range of urban, suburban, and rural settings is an encouraging indicator that the underlying image segmentation approach is not narrowly optimized to any particular streetscape type. Across the 18 AOIs, street-view panoramas were available for 73.4% of structures, and 49.0% yielded successful LFE/HDSL extraction. Addressing this bottleneck through multi-angle image acquisition, supplementary imagery providers, or alternative visible elevation indicators such as garage thresholds and foundation steps (as recommended by Ho et al., 2024) represents the most direct path to improving regional LFE coverage. Weaker performance in Harris County Pocket 1 ($R^2$_CV = 0.159) reflects the challenges of a large, heterogeneous urban AOI with multiple overlapping construction eras and wide HDSL variance that the terrain and hydrologic predictor set does not fully capture; however, the near-zero overfitting gap (Gap = 0.024) indicates that while the model explains only a modest fraction of total variance, its predictions are consistent and stable across cross-validation folds rather than artifacts of overfitting to the training data.

The explicit exclusion of five AOIs where imputation quality was insufficient is a methodological contribution in itself. Many published imputation studies apply models uniformly without performance-gated exclusion, potentially introducing unreliable estimates for a subset of the housing stock. By reporting model performance per AOI and applying a quality threshold, this study provides a transparent framework for communicating the geographic limits of ML imputation reliability. Future work should investigate whether transfer learning, by training on pooled data from high-quality AOIs and applying to data-sparse ones, can extend imputation coverage to currently excluded areas.

Several sources of uncertainty compound in the integrated pipeline and should be acknowledged when interpreting the results. Uncertainty arises from the USACE DDF itself, which was calibrated on Gulf Coast claims data dominated by single-story slab-on-grade construction and may not accurately represent damage in rural AOIs where mobile homes and pier-and-beam structures are more prevalent. Market value-based damage estimates also abstract over substantial within-AOI heterogeneity in structural quality, building materials, and contents. Taken together, these uncertainties suggest the damage estimates are best interpreted as order-of-magnitude regional screening values rather than precise property-level predictions.

The findings should be interpreted within the scope of a regional screening framework rather than as a replacement for engineering-grade elevation surveys. Direct extraction remains constrained by the availability and quality of street-view imagery, by vegetation and setback conditions, and by whether architectural features revealing floor height are visible from the public right-of-way. The imputation stage, while often effective, is not uniformly transferable across AOIs and depends on the quantity and quality of the directly extracted observations available for training. Downstream loss estimates also inherit uncertainty from the elevation pipeline, the flood surfaces, and the depth-damage functions used to convert inundation into economic loss. These limitations define the current operating envelope of the method rather than diminish its contribution. Future research should expand direct coverage through multi-angle or multi-provider imagery, test pooled and transfer-learning strategies for data-sparse AOIs, evaluate structure-type-specific imputation and damage functions, and validate modeled losses against claims or observed post-event damage to strengthen the pathway from regional screening to operational deployment.

A practical implementation pathway begins by assembling parcel geometries, address records, property values, and hazard surfaces for the target jurisdiction, followed by parcel-level retrieval and quality screening of street-view imagery. Direct LFE/HDSL extraction should then be completed first so that implementation teams can quantify image availability, facade visibility, and extraction yield before attempting to fill data gaps. Machine-learning imputation should be applied only after AOI-level diagnostics confirm that sufficient training observations exist and that cross-validated performance exceeds a predefined reliability threshold; where those conditions are not met, the output should remain explicitly partial rather than be forced to full coverage. The resulting dataset can then be integrated with flood surfaces and depth-damage functions to produce maps and summaries for mitigation prioritization, emergency planning, insurance screening, or floodplain management. Key stakeholders include local floodplain managers, emergency managers, appraisal districts, and insurers, while the main barriers to adoption are imagery incompleteness, validation requirements, and the need for routine data integration. Those barriers are manageable if implementation is staged, performance thresholds are explicit, and the database is updated as imagery and parcel records change over time.

The pipeline has practical applications across several domains of flood risk governance. For emergency management, the FDIS maps identify which neighborhoods and individual blocks face the greatest interior flooding hazard under a given scenario, supporting pre-event resource positioning and evacuation priority designations. For floodplain management and community resilience planning, the property-level damage estimates provide a basis for ranking mitigation investments (buyouts, structural elevations, or floodway improvements) by cost-effectiveness across a community's housing stock. For the insurance sector, the extracted and imputed HDSL data offer an evidence base for refining NFIP risk classifications for the large proportion of policyholders who lack valid Elevation Certificates, a particularly pressing need in Texas given the state's flood risk exposure and historically low EC coverage. The properties currently lacking elevation data are not randomly distributed geographically and excluding them from damage estimation likely underestimates community-scale risk in systematically vulnerable areas, an effect that will be quantified through sensitivity analysis in the RQ3 output.

## 7. Conclusion

This study presents a scalable, three-stage pipeline for property-level flood risk assessment grounded in lowest floor elevation data extracted from publicly available street-view imagery and extended through machine learning imputation. Building on the Elev-Vision framework of Ho et al.(2025; 2024), LFE and HDSL extraction was deployed across 18 AOIs in Texas, achieving street-view imagery coverage of 73.4% across 12,241 total structures and successful LFE/HDSL extraction for 49.0% of structures (5,992 properties), representing the first regional-scale application of this approach. A two-tier ML imputation framework employing Random Forest and Gradient Boosting regressors, trained on 16 spatial, terrain, hydrologic, and flood exposure features and systematically evaluated through 5-fold cross-validation, achieved $R^2_CV$ values of 0.159 to 0.974 across the 13 qualifying AOIs (RMSE = 0.035–0.277 m), with five AOIs explicitly excluded where training data quality was insufficient to support reliable imputation. Integration of the combined LFE dataset with Fathom 1-in-100 year flood surfaces produces property-level FDIS estimates and depth-damage-based economic loss calculations for all 18 AOIs.

Three contributions stand out in our study. First, this work establishes that the Elev-Vision extraction pipeline maintains its accuracy when scaled from a single neighborhood to a geographically diverse multi-community deployment, a necessary precondition for any regional application of street-view-based LFE mapping. Second, the study introduces a machine learning HDSL imputation approach grounded in physically motivated hydrologic terrain predictors, with performance-gated exclusion that explicitly communicates where imputation is and is not reliable, a methodological standard that distinguishes this framework from prior imputation studies in the flood risk literature. Third, to the authors' knowledge, this is the first multi-community, property-level flood damage assessment in Texas that integrates both extracted and ML-imputed LFE data, providing a substantive regional risk characterization alongside the methodological advances.

Several directions merit future investigation. Improving front-door visibility through multi-angle panorama acquisition and alternative imagery providers represents the most direct path to increasing direct extraction coverage. Transfer learning from data-rich AOIs to imputation-excluded communities could extend ML-based HDSL estimation to areas currently beyond the reach of the approach. Validation of modeled damage estimates against NFIP claims records from major Texas flood events would provide empirical grounding for the damage function parameters and their applicability to the Texas built environment. Finally, applying the pipeline to climate-adjusted flood scenarios would enable long-range risk characterization and support adaptation planning in communities whose flood exposure is projected to intensify over coming decades. AOI-level results, maps, and model performance diagnostics are available at Zenodo: zenodo_xiangpengli_HDSL.

## Acknowledgments

The authors acknowledge funding support from

## Data Availability

Street-view imagery data are accessible via the Google Street View Static API. Fathom flood model tiles are available under research license from Fathom (https://www.fathom.global). Terrain and hydrologic rasters (HAND, D2Stream) are publicly available from the USGS and NOAA. County parcel and appraisal data are available from respective county appraisal district open data portals. Processed LFE extraction and imputation outputs are available from the corresponding author upon reasonable request.

## Conflict of Interest

The authors declare no conflicts of interest.

## References

Amadio, M., Mysiak, J., Carrera, L., & Koks, E. (2016). Improving flood damage assessment models in Italy. *Natural Hazards*, *82*(3), 2075–2088. https://doi.org/10.1007/s11069-016-2286-0

Bates, P. D., Quinn, N., Sampson, C., Smith, A., Wing, O., Sosa, J., Savage, J., Olcese, G., Neal, J., Schumann, G., Giustarini, L., Coxon, G., Porter, J. R., Amodeo, M. F., Chu, Z., Lewis-Gruss, S., Freeman, N. B., Houser, T., Delgado, M., … Krajewski, W. F. (2021). Combined Modeling of US Fluvial, Pluvial, and Coastal Flood Hazard Under Current and Future Climates. *Water Resources Research*, *57*(2), e2020WR028673. https://doi.org/10.1029/2020WR028673

Biljecki, F., & Ito, K. (2021). Street view imagery in urban analytics and GIS: A review. *Landscape and Urban Planning*, *215*, 104217. https://doi.org/10.1016/j.landurbplan.2021.104217

Bodoque, J., Guardiola-Albert, C., Aroca-Jiménez, E., Eguibar, M., & Martínez-Chenoll, M. (2016). Flood Damage Analysis: First Floor Elevation Uncertainty Resulting from LiDAR-Derived Digital Surface Models. *Remote Sensing*, *8*(7), 604. https://doi.org/10.3390/rs8070604

Bonczak, B., & Kontokosta, C. E. (2019). Large-scale parameterization of 3D building morphology in complex urban landscapes using aerial LiDAR and city administrative data. *Computers, Environment and Urban Systems*, *73*, 126–142. https://doi.org/10.1016/j.compenvurbsys.2018.09.004

Breiman, L. (2001). Random Forests. *Machine Learning*, *45*(1), 5–32. https://doi.org/10.1023/A:1010933404324

Chen, F.-C., Jahanshahi, M. R., Johnson, D., & Delp, E. J. (2019a, November 15). Vision-based Decision Support for Flood Risk Assessment Using Google Street View Images. *Structural Health Monitoring 2019*. Structural Health Monitoring 2019. https://doi.org/10.12783/shm2019/32472

Chen, F.-C., Jahanshahi, M. R., Johnson, D., & Delp, E. J. (2019b, November 15). Vision-based Decision Support for Flood Risk Assessment Using Google Street View Images. *Structural Health Monitoring 2019*. Structural Health Monitoring 2019. https://doi.org/10.12783/shm2019/32472

Chen, F.-C., Subedi, A., Jahanshahi, M. R., Johnson, D. R., & Delp, E. J. (2022). Deep Learning–Based Building Attribute Estimation from Google Street View Images for Flood Risk Assessment Using Feature Fusion and Task Relation Encoding. *Journal of Computing in Civil Engineering*, *36*(6), 04022031. https://doi.org/10.1061/(ASCE)CP.1943-5487.0001025

Diaz, N. D., Highfield, W. E., Brody, S. D., & Fortenberry, B. R. (2022a). Deriving First Floor Elevations within Residential Communities Located in Galveston Using UAS Based Data. *Drones*, *6*(4), 81. https://doi.org/10.3390/drones6040081

Diaz, N. D., Highfield, W. E., Brody, S. D., & Fortenberry, B. R. (2022b). Deriving First Floor Elevations within Residential Communities Located in Galveston Using UAS Based Data. *Drones*, *6*(4), 81. https://doi.org/10.3390/drones6040081

Esparza, M., Ho, Y.-H., Brody, S., & Mostafavi, A. (2025). Improving flood damage estimation by integrating property elevation data. *International Journal of Disaster Risk Reduction*, *118*, 105251. https://doi.org/10.1016/j.ijdrr.2025.105251

Fan, Z., Zhang, F., Loo, B. P. Y., & Ratti, C. (2023). Urban visual intelligence: Uncovering hidden city profiles with street view images. *Proceedings of the National Academy of Sciences*, *120*(27), e2220417120. https://doi.org/10.1073/pnas.2220417120

Feng, Y., Xiao, Q., Brenner, C., Peche, A., Yang, J., Feuerhake, U., & Sester, M. (2022). Determination of building flood risk maps from LiDAR mobile mapping data. *Computers, Environment and Urban Systems*, *93*, 101759. https://doi.org/10.1016/j.compenvurbsys.2022.101759

Fong, C., & Tyler, M. (2021). Machine Learning Predictions as Regression Covariates. *Political Analysis*, *29*(4), 467–484. https://doi.org/10.1017/pan.2020.38

Friedman, J. H. (2001). Greedy function approximation: A gradient boosting machine. *The Annals of Statistics*, *29*(5). https://doi.org/10.1214/aos/1013203451

Gems, B., Mazzorana, B., Hofer, T., Sturm, M., Gabl, R., & Aufleger, M. (2016). 3-D hydrodynamic modelling of flood impacts on a building and indoor flooding processes. *Natural Hazards and Earth System Sciences*, *16*(6), 1351–1368. https://doi.org/10.5194/nhess-16-1351-2016

Guo, M., Gong, J., & Whytlaw, J. L. (2022). Large-scale cloud-based building elevation data extraction and flood insurance estimation to support floodplain management. *International Journal of Disaster Risk Reduction*, *69*, 102741. https://doi.org/10.1016/j.ijdrr.2021.102741

Hengl, T., Nussbaum, M., Wright, M. N., Heuvelink, G. B. M., & Gräler, B. (2018). Random forest as a generic framework for predictive modeling of spatial and spatio-temporal variables. *PeerJ*, *6*, e5518. https://doi.org/10.7717/peerj.5518

Highfield, W. E., Peacock, W. G., & Van Zandt, S. (2014). Mitigation Planning: Why Hazard Exposure, Structural Vulnerability, and Social Vulnerability Matter. *Journal of Planning Education and Research*, *34*(3), 287–300. https://doi.org/10.1177/0739456X14531828

Ho, Y., Li, L., & Mostafavi, A. (2025). Integrated vision language and foundation model for automated estimation of building lowest floor elevation. *Computer-Aided Civil and Infrastructure Engineering*, *40*(1), 75–90. https://doi.org/10.1111/mice.13310

Ho, Y.-H., Lee, C.-C., Diaz, N., Brody, S., & Mostafavi, A. (2024). ELEV-VISION: Automated Lowest Floor Elevation Estimation from Segmenting Street View Images. *ACM Journal on Computing and Sustainable Societies*, *2*(2), 1–18. https://doi.org/10.1145/3661832

Huang, Y., Zhang, F., Gao, Y., Tu, W., Duarte, F., Ratti, C., Guo, D., & Liu, Y. (2023). Comprehensive urban space representation with varying numbers of street-level images. *Computers, Environment and Urban Systems*, *106*, 102043. https://doi.org/10.1016/j.compenvurbsys.2023.102043

Ibebuchi, C. C., & Abu, I.-O. (2025). Probabilistic flood susceptibility mapping using explainable AI for the Western United States. *Environmental Research Communications*, *7*(10), 105008. https://doi.org/10.1088/2515-7620/ae0c5c

Johnson, D. R. (2019). Improved Methods for Estimating Flood Depth Exceedances Within Storm Surge Protection Systems. *Risk Analysis*, *39*(4), 890–905. https://doi.org/10.1111/risa.13213

Kirillov, A., Mintun, E., Ravi, N., Mao, H., Rolland, C., Gustafson, L., Xiao, T., Whitehead, S., Berg, A. C., Lo, W.-Y., Dollár, P., & Girshick, R. (2023). *Segment Anything* (Version 1). arXiv. https://doi.org/10.48550/ARXIV.2304.02643

Li, X., Ma, J., Li, B., & Ali, M. (2026). Quantifying the social costs of power outages and restoration disparities across four U.S. hurricanes. *International Journal of Disaster Risk Reduction*, *136*, 106086. https://doi.org/10.1016/j.ijdrr.2026.106086

Lu, X., & Li, Z. (2024). Spatial analysis and machine learning: Towards integrated predictive modeling advancements. *Theoretical and Natural Science*, *36*(1), 152–157. https://doi.org/10.54254/2753-8818/36/20240538

Merz, B., Kreibich, H., Schwarze, R., & Thieken, A. (2010). Review article "Assessment of economic flood damage" *Natural Hazards and Earth System Sciences*, *10*(8), 1697–1724. https://doi.org/10.5194/nhess-10-1697-2010

Nederhoff, K., Crosby, S. C., Van Arendonk, N. R., Grossman, E. E., Tehranirad, B., Leijnse, T., Klessens, W., & Barnard, P. L. (2024). Dynamic Modeling of Coastal Compound Flooding Hazards Due to Tides, Extratropical Storms, Waves, and Sea-Level Rise: A Case Study in the Salish Sea, Washington (USA). *Water*, *16*(2), 346. https://doi.org/10.3390/w16020346

Ning, H., Li, Z., Ye, X., Wang, S., Wang, W., & Huang, X. (2022). Exploring the vertical dimension of street view image based on deep learning: A case study on lowest floor elevation estimation. *International Journal of Geographical Information Science*, *36*(7), 1317–1342. https://doi.org/10.1080/13658816.2021.1981334

proog128. (n.d.). *GSVPanoDepth.js* [Computer software]. Retrieved https://github.com/proog128/GSVPanoDepth.js/tree/master

Radford, A., Kim, J. W., Hallacy, C., Ramesh, A., Goh, G., Agarwal, S., Sastry, G., Askell, A., Mishkin, P., Clark, J., Krueger, G., & Sutskever, I. (2021). Learning Transferable Visual Models From Natural Language Supervision. In M. Meila & T. Zhang (Eds.),

*Proceedings of the 38th International Conference on Machine Learning* (Vol. 139, pp. 8748–8763). PMLR. https://proceedings.mlr.press/v139/radford21a.html

Ronco, P., Gallina, V., Torresan, S., Zabeo, A., Semenzin, E., Critto, A., & Marcomini, A. (2014). The KULTURisk Regional Risk Assessment methodology for water-related natural hazards – Part 1: Physical–environmental assessment. *Hydrology and Earth System Sciences*, *18*(12), 5399–5414. https://doi.org/10.5194/hess-18-5399-2014

Scawthorn, C., Blais, N., Seligson, H., Tate, E., Mifflin, E., Thomas, W., Murphy, J., & Jones, C. (2006). HAZUS-MH Flood Loss Estimation Methodology. I: Overview and Flood Hazard Characterization. *Natural Hazards Review*, *7*(2), 60–71. https://doi.org/10.1061/(ASCE)1527-6988(2006)7:2(60)

Sebastian, A., Bader, D. J., Nederhoff, C. M., Leijnse, T. W. B., Bricker, J. D., & Aarninkhof, S. G. J. (2021). Hindcast of pluvial, fluvial, and coastal flood damage in Houston, Texas during Hurricane Harvey (2017) using SFINCS. *Natural Hazards*, *109*(3), 2343–2362. https://doi.org/10.1007/s11069-021-04922-3

Shultz, J. M., & Galea, S. (2017). Mitigating the Mental and Physical Health Consequences of Hurricane Harvey. *JAMA*, *318*(15), 1437. https://doi.org/10.1001/jama.2017.14618

Taghinezhad, A., Friedland, C. J., Rohli, R. V., & Marx, B. D. (2020). An Imputation of First-Floor Elevation Data for the Avoided Loss Analysis of Flood-Mitigated Single-Family Homes in Louisiana, United States. *Frontiers in Built Environment*, *6*, 138. https://doi.org/10.3389/fbuil.2020.00138

Tonn, G., & Czajkowski, J. (2022). US tropical cyclone flood risk: Storm surge versus freshwater. *Risk Analysis*, *42*(12), 2748–2764. https://doi.org/10.1111/risa.13890

Touzani, S., & Granderson, J. (2021). Open Data and Deep Semantic Segmentation for Automated Extraction of Building Footprints. *Remote Sensing*, *13*(13), 2578. https://doi.org/10.3390/rs13132578

U.S. Army Corps of Engineers (USACE). (2003). *Economic Guidance Memorandum (EGM) 04-01: Generic Depth-Damage Relationships for Residential Structures*. https://planning.erdc.dren.mil/toolbox/library/EGMs/egm04-01.pdf

Wagenaar, D., Lüdtke, S., Schröter, K., Bouwer, L. M., & Kreibich, H. (2018). Regional and Temporal Transferability of Multivariable Flood Damage Models. *Water Resources Research*, *54*(5), 3688–3703. https://doi.org/10.1029/2017WR022233

Wing, O. E. J., Bates, P. D., Sampson, C. C., Smith, A. M., Johnson, K. A., & Erickson, T. A. (2017). Validation of a 30 m resolution flood hazard model of the conterminous U nited S tates. *Water Resources Research*, *53*(9), 7968–7986. https://doi.org/10.1002/2017WR020917

Wing, O. E. J., Bates, P. D., Smith, A. M., Sampson, C. C., Johnson, K. A., Fargione, J., & Morefield, P. (2018). Estimates of present and future flood risk in the conterminous United States. *Environmental Research Letters*, *13*(3), 034023. https://doi.org/10.1088/1748-9326/aaac65

Wu, M., Zeng, W., & Fu, C.-W. (2021). FloorLevel-Net: Recognizing Floor-Level Lines With Height-Attention-Guided Multi-Task Learning. *IEEE Transactions on Image Processing*, *30*, 6686–6699. https://doi.org/10.1109/TIP.2021.3096090

Yildirim, E., Just, C., & Demir, I. (2022). Flood risk assessment and quantification at the community and property level in the State of Iowa. *International Journal of Disaster Risk Reduction*, *77*, 103106. https://doi.org/10.1016/j.ijdrr.2022.103106

Yuan, Q., & Xia, B. (2024). Cross-level and multiscale CNN-Transformer network for automatic building extraction from remote sensing imagery. *International Journal of Remote Sensing*, *45*(9), 2893–2914. https://doi.org/10.1080/01431161.2024.2339199

Zarekarizi, M., Srikrishnan, V., & Keller, K. (2020). Neglecting uncertainties biases house-elevation decisions to manage riverine flood risks. *Nature Communications*, *11*(1), 5361. https://doi.org/10.1038/s41467-020-19188-9

Zhang, X., Kang, A., Ye, M., Song, Q., Lei, X., & Wang, H. (2023). Influence of Terrain Factors on Urban Pluvial Flooding Characteristics: A Case Study of a Small Watershed in Guangzhou, China. *Water*, *15*(12), 2261. https://doi.org/10.3390/w15122261